\documentclass[runningheads]{llncs}

\usepackage[table]{xcolor}
\usepackage{times}
\usepackage{epsfig}
\usepackage{graphicx}
\usepackage{amsmath}
\usepackage{amssymb}
\usepackage{xcolor}
\usepackage{tabularx}
\usepackage{graphicx}

\usepackage{xcolor}
\usepackage{url}
\usepackage{algorithm}
\usepackage[noend]{algpseudocode}
\usepackage{algorithmicx}

\usepackage{microtype}
\usepackage{lipsum}
\usepackage{graphicx}
\usepackage{subcaption}
\usepackage{sidecap}
\usepackage{caption}
\usepackage{booktabs} %
\usepackage{amsfonts}       %
\usepackage{nicefrac}       %
\usepackage{microtype}      %
\usepackage{comment}
\usepackage{enumitem}
\usepackage{wrapfig,tikz}
\usepackage{amsmath,longtable,fancyhdr}
\usepackage{bigstrut, tabularx, multirow, makecell, diagbox}
\usepackage[export]{adjustbox}
\usepackage{graphicx,float}

\usepackage{stackrel}
\usepackage{color}
\usepackage{colortbl}
\usepackage{theoremref}
\usepackage{cases}
\usepackage{mathabx}
\usepackage{dsfont}

\newcolumntype{C}[1]{>{\centering\arraybackslash}m{#1}}
\newcolumntype{P}[1]{>{\centering\arraybackslash}p{#1}}

\newcommand{\mcal}{\mathcal}

\usepackage{mdframed}

\newcommand{\be}{\begin{equation}}
\newcommand{\ee}{\end{equation}}
\definecolor{Gray}{gray}{0.85}
\definecolor{LightCyan}{rgb}{0.88,1,1}

\makeatletter
\usepackage{xspace} 
\def\@onedot{\ifx\@let@token.\else.\null\fi\xspace}
\DeclareRobustCommand\onedot{\futurelet\@let@token\@onedot}

\newcommand{\bfx}{\mathbf{x}}

\newcommand{\bfI}{\mathbf{I}}

\newcommand{\bfzero}{\mathbf{0}}

\newcommand{\bftheta}{{\boldsymbol{\theta}}}

\newcommand{\bfs}{\mathbf{s}}

\definecolor{blue1}{RGB}{0,128,255}
\definecolor{blue3}{RGB}{0,0,128}
\definecolor{darkpastelgreen}{rgb}{0.01, 0.75, 0.24}
\definecolor{cerulean}{rgb}{0.0, 0.48, 0.65}

\def\eg{\emph{e.g}\onedot}

\def\ie{\emph{i.e}\onedot}

\def\aka{a.k.a\onedot}

\def\etal{\emph{et al}\onedot}

\usepackage{amsmath,amsfonts,bm}

\def\eqref#1{equation~\ref{#1}}

\def\1{\bm{1}}

\def\eps{{\epsilon}}

\DeclareMathAlphabet{\mathsfit}{\encodingdefault}{\sfdefault}{m}{sl}
\SetMathAlphabet{\mathsfit}{bold}{\encodingdefault}{\sfdefault}{bx}{n}

\usepackage{bm}
\usepackage{float}

\usepackage{xspace}

\makeatletter
\DeclareRobustCommand\onedot{\futurelet\@let@token\@onedot}
\def\@onedot{\ifx\@let@token.\else.\null\fi\xspace}

\def\eg{\emph{e.g}\onedot} 
\def\ie{\emph{i.e}\onedot}

\def\etal{\emph{et al}\onedot}
\makeatother

    \makeatletter
\def\@fnsymbol#1{\ensuremath{\ifcase#1\or * \or * \or
   \mathsection\or \mathparagraph\or \|\or **\or \dagger\dagger
   \or \ddagger\ddagger \else\@ctrerr\fi}}
    \makeatother

\makeatletter
\DeclareRobustCommand\onedot{\futurelet\@let@token\@onedot}
\def\@onedot{\ifx\@let@token.\else.\null\fi\xspace}

\def\eg{\emph{e.g}\onedot} 
\def\ie{\emph{i.e}\onedot}

\def\etal{\emph{et al}\onedot}
\makeatother

    \makeatletter
\def\@fnsymbol#1{\ensuremath{\ifcase#1\or * \or * \or
   \mathsection\or \mathparagraph\or \|\or **\or \dagger\dagger
   \or \ddagger\ddagger \else\@ctrerr\fi}}
    \makeatother

\usepackage{eccvabbrv}

\usepackage{graphicx}
\usepackage{booktabs}

\usepackage[accsupp]{axessibility}  

\usepackage[pagebackref,breaklinks,colorlinks,citecolor=eccvblue]{hyperref}
\hypersetup{
  colorlinks,
  linkcolor={red!50!black},
  citecolor={blue!50!black},
  urlcolor={blue!80!black}
  }
  \definecolor{orange}{HTML}{ff7f0e}
  \definecolor{blue}{HTML}{1f77b4}

\usepackage[capitalise]{cleveref}
\usepackage{orcidlink}

\begin{document}

\title{Layered Rendering Diffusion Model for Controllable Zero-Shot Image Synthesis} 

\titlerunning{Abbreviated paper title}

\author{Zipeng Qi$^{\ast}$\inst{1} \and Guoxi Huang$^{\ast \dagger}$\inst{2}
 \and Chenyang Liu \inst{1} \and Fei Ye\inst{3}}

\authorrunning{Z. Qi~et al.}

\institute{Beihang University \and
University of Bristol \and
Mohamed bin Zayed University of Artificial Intelligence }

\def\thefootnote{$\ast$}\footnotetext{Equal Contribution: Z. Qi and G. Huang}
\def\thefootnote{$\dagger$}\footnotetext{Corresponding author: G. Huang (guoxi.huang@bristol.ac.uk)}
\def\thefootnote{\arabic{footnote}}


\maketitle
\begin{abstract}
This paper introduces innovative solutions to enhance spatial controllability in diffusion models reliant on text queries. We first introduce vision guidance as a foundational spatial cue within the perturbed distribution.
This significantly refines the search space in a zero-shot paradigm to focus on the image sampling process adhering to the spatial layout conditions. To precisely control the spatial layouts of multiple visual concepts with the employment of vision guidance, we propose a universal framework, \textbf{L}ayered \textbf{R}endering \textbf{Diff}usion (LRDiff), which constructs an image-rendering process with multiple layers, each of which applies the vision guidance to instructively estimate the denoising direction for a single object. Such a layered rendering strategy effectively prevents issues like unintended conceptual blending or mismatches while allowing for more coherent and contextually accurate image synthesis. The proposed method offers a more efficient and accurate means of synthesising images that align with specific layout and contextual requirements. Through experiments, we demonstrate that our method outperforms existing techniques, both quantitatively and qualitatively, in two specific layout-to-image tasks: bounding box-to-image and instance mask-to-image. Furthermore, we extend the proposed framework to enable spatially controllable editing. 
The project page is available \href{https://qizipeng.github.io/LRDiff_projectPage}{here}.

\end{abstract}

\section{Introduction}
\label{intro}
Large-scale Text-to-Image (T2I) diffusion models trained at scale (e.g., 250 million captioned images for DALL$\cdot$E~\cite{ramesh2021zero}) have recently shown remarkable capabilities in generating high-fidelity images, covering diverse concepts. Meanwhile, the excellent data synthesis capabilities of diffusion models have been extensively leveraged in diverse fields, encompassing 3D modelling~\cite{gu2023nerfdiff,wynn2023diffusionerf}, training data creation~\cite{wu2023datasetdm}, video generation~\cite{ho2022video}, among others. 

Despite the versatility of text, diffusion models relying solely on text input encounter challenges in achieving spatial controllability. This hinders fine control over the layout of generated results. Existing methods to tackle this issue mainly fall into two categories: (1) Inputting additional spatial layout entities (e.g. semantic maps~\cite{zhang2023adding} or serialised bounding boxes~\cite{li2023gligen}) and extra parameterised components through fine-tuning models; (2) Manipulating the attention map through gradient computation aims to enhance the attention score of noise features and text within specific areas~\cite{balaji2022ediffi,couairon2023zero}. 
Although the former can achieve competitive results with precise spatial alignment, it is noteworthy that they incur substantial computational costs for fine-tuning models and labour costs for data curation. The latter modifies all features simultaneously, presenting a challenge in distinguishing adjacent objects with the same category (see the giraffe example in Fig.~\ref{box_results_qualitative}). Besides, the latter, which directly updates the attention map through gradients, will increase the latency due to frequent backpropagation.

In this paper, our focus is on achieving controllable image synthesis without model re-training or fine-tuning. We propose a universal framework, a two-stage Layered Rendering Diffusion (LRDiff), specifically designed for the above task in a zero-shot paradigm. LRDiff aims to process multiple visual concepts without blending their representations and aligns the results with the input spatial conditions, such as bounding boxes or instant masks. The denoising process in LRDiff is divided into two separate sections. In the first section, we estimate the denoising direction of each object in layers to ensure the accuracy of layouts, employing an innovative concept termed `vision guidance'. The second denoising section focuses on enriching the texture details and aligning the high-level concepts, guided by the global context of the original caption. Vision guidance, constituting one of the cores of LRDiff, provides a spatial cue for explicitly estimating the denoising direction of each object in layers to ensure the accuracy of its location, shapes, or contour without gradient computation, as illustrated in Fig.~\ref{fig:pipline}(a). The implementation of vision guidance empowers LRDiff with zero-shot capabilities for each object and allows adaptation to two common controllable image synthesis tasks, including box-to-image and mask-to-image.

Our experimental results demonstrate that the proposed LRDiff provides excellent spatial controllability for T2I diffusion models while generating photorealistic scene images.
Compared to previous methods, including BoxDiff~\cite{xie2023boxdiff}, DenseDiffusion~\cite{kim2023dense}, and Paint-with-Words (eDiffi-Pww)~\cite{balaji2022ediffi}, among others, our results show improved performance, both quantitatively and qualitatively, as demonstrated in Fig.~\ref{box_results_qualitative}
and Fig.~\ref{mask_results_qualitative}.
The main contributions of this paper are summarised as follows:
\begin{itemize}
    \item We introduce a universal framework for controllable image synthesis, which is a two-stage layered rendering diffusion model to process multiple visual concepts in layers while aligning the results with the global text.
    \item We propose visual guidance, independent of the network structure, and incorporate it into each layer to achieve spatial controllability for each object. Vision guidance provides a spatial cue in a zero-shot paradigm without the need for backpropagation.

    \item Three applications are enabled by the proposed framework: bounding box-to-image, instance mask-to-image, and controllable image editing.
\end{itemize} 

\section{Related Works}
\label{related works}
\noindent \textbf{Text-to-Image (T2I) Models.}
To adhere to some specifications described by free-form text in image generation, T2I typically models the image distribution along with the encoded latent embeddings of the text prompts as the condition entities via pre-trained language models, such as CLIP~\cite{radford2021learning}.
Large-scale text-to-image models can be categorised as auto-regressive models~\cite{ramesh2021zero,ding2021cogview,gafni2022make,yu2022scaling} and diffusion-based models~\cite{nichol2021glide,rombach2022high,ramesh2022hierarchical,saharia2022photorealistic}.
Inspired by non-equilibrium statistical physics, Dickstein~\etal~\cite{sohl2015deep} pioneeringly introduced the diffusion model, the concept of which is revisited in Sec.~\ref{sec:preliminaries}.
To accelerate training and sampling speed, the latent diffusion model~\cite{rombach2022high}, \aka Stable Diffusion (SD) is developed to operate the diffusion process in the latent space~\cite{esser2021taming} instead of the pixel space.
However, when it comes to the intricate spatial semantic arrangement of multiple objects within a scene, T2I diffusion models fall short, exhibiting object leakage and a lack of awareness regarding spatial dependencies.

\vspace{0.15cm}
\noindent \textbf{Layout-to-Image (L2I) Diffusion Models.}
Current layout-guided generation methods can be broadly classified into two main categories based on the necessity of a training process: (1) methods that involve fine-tuning diffusion models~\cite{li2023gligen,yang2023reco,zhang2023adding,chen2023integrating,cheng2023layoutdiffuse,mou2023t2i,yang2023paint,qin2023unicontrol,zhao2023uni,kim2023diffblender,ju2023human}, and (2) training-free approaches~\cite{couairon2023zero,qu2023layoutllm,xie2023boxdiff,he2023localized,bar2023multidiffusion,kim2023dense,zeng2023scenecomposer,phung2023grounded,zhang2023controllable,ma2023directed,zheng2023layoutdiffusion,yu2023freedom,bansal2023universal,li2023drivingdiffusion}. The former achieves locality-awareness by incorporating layout information as an additional condition to the pre-trained T2I diffusion model. 
Methods necessitating fine-tuning, like ControlNet\cite{zhang2023adding}, GLIGEN~\cite{li2023gligen} and T2I-Adaptor~\cite{mou2023t2i} integrate extra modules into the backbone network. These modules work in concert with spatial control entities to ensure the generated images match the specified spatial conditions.
ReCo~\cite{yang2023reco} and GeoDiffusion~\cite{chen2023integrating} augment the textual tokens by incorporating new positional tokens arranged in sequences akin to short natural language sentences.
However, it's worth noting that these approaches require further training on curated datasets collected with paired annotations, which imposes significant computational and labour costs, bottlenecking applications in an open world.
On the other hand, the second group of methods, such as Paint-with-words (eDiffi-Pww)~\cite{balaji2022ediffi} and ZestGuide~\cite{couairon2023zero}, endows T2I diffusion models with localisation abilities through manipulating the cross-attention maps in the estimator network, amplifying the attention score for the text tokens that specify an object. However, when the initial noise does not tend to generate the target objects, it is difficult to change the direction of denoising by only manipulating the attention map.
Besides, these methods are likely to cause the blending of appearances of adjacent objects sharing the same visual concept. This issue is exemplified in the first and second columns of Fig.~\ref{box_results_qualitative}, where the phenomenon is clearly observable. 

\vspace{0.15cm}
\noindent \textbf{Image Editing with Diffusion Models.} 
Image editing, as a fundamental task in computer graphics, can be achieved by modifying a real image by inputting auxiliary entities, including scribble~\cite{meng2021sdedit}, mask~\cite{avrahami2022blended}, or reference image~\cite{brooks2023instructpix2pix}.
Recent models for text-conditioned image editing~\cite{hertz2023prompttoprompt,kawar2023imagic,meng2021sdedit,couairon2022diffedit} harness CLIP~\cite{radford2021learning} embedding guidance combined with pre-trained T2I diffusion models, achieving excellent results across a range of editing tasks.
Research in this field primarily advances along three directions: (1) zero-shot algorithms that steer the denoising process towards a desired CLIP embedding direction by manipulating the attention maps of the cross-attention mechanisms~\cite{hertz2022prompt,parmar2023zero,meng2021sdedit,mokady2023null,kim2022diffusionclip}; (2) textual token vector optimisation ~\cite{ruiz2023dreambooth,gal2022image}; (3) fine-tuning T2I diffusion models on curated datasets with matched annotations~\cite{brooks2023instructpix2pix,nguyen2023visual}.
In contrast to prior works that rely on text prompts to guide the editing, we aim to leverage the proposed vision guidance information to better assist the generation process.

\section{Preliminaries}
\label{sec:preliminaries}
\noindent \textbf{Denoising Diffusion Probabilistic Models (DDPMs).}
DDPM~\cite{ho2020denoising} involves a forward-time diffusion process and a reverse-time denoising process from a prior distribution. 
Let $\bfx_0 \in \mathbb{R}^{h\times w \times D}$ be a sample from the data distribution $p_0(\bfx)$. When using a total of $T$ noise scales in the forward-time diffusion process, the discrete Markov chain is
\begin{align}
    \bfx_t = \sqrt{1-\beta_{t}} \bfx_{t-1} + \sqrt{\beta_{t}} \eps_{t-1}, \quad t=1,\cdots,T ~~~~, \label{eqn:ddpm}
\end{align}
where $\eps_t$ denotes the noise sampled from $\mcal{N}\big( \bfzero, \bfI \big)$ at timestep $t$, $\{\beta_t\}_{t=1}^{T}$ is a pre-defined variance schedule.
When recursively applying the noise perturbations \cref{eqn:ddpm} to a real sample $\bfx_0 \sim p_0(\bfx)$, it ends up with $\bfx_T \sim \mcal{N}\big( \bfzero, \bfI \big)$. The reverse-time denoising process can be defined as:
\begin{equation}
    \begin{aligned}
        \bfx_{t-1} &= \tilde{\alpha}_t \bfx_{t} + \underbrace{\tilde{\beta}_t \nabla_\bfx \log p_t(\bfx)}_{\text{direction pointing to } \bfx_t} + \underbrace{\sigma_t \eps_t}_{\text{random noise}},\\
         &\approx  \tilde{\alpha}_t \bfx_{t} + \underbrace{\tilde{\beta}_t \hat{\bfs}_t}_{\text{estimated direction pointing to } \bfx_t} + \underbrace{\sigma_t \eps_t}_{\text{random noise}}.
        \label{eqn:backward_dm}
    \end{aligned}
\end{equation}
where $\tilde{\alpha}_t$, $\tilde{\beta}_t$, and $\sigma_t$ denote the coefficients, the values of which can be derived from $\beta$.
Practically, the score of the perturbed data distribution, $\nabla_\bfx \log p_t(\bfx)$ for all $t$, can be estimated with a score network $\bfs_\bftheta(\bfx_{t}, t)$ optimised by using score matching~\cite{hyvarinen2005estimation,song2019generative}.
After training to get the optimal solution $ \hat{\bfs}_t = \bfs_\bftheta^*(\bfx_{t}, t) \approx \nabla_\bfx \log p_t(\bfx)$, new samples can be generated by starting from $\bfx_T \sim \mcal{N}\big( \bfzero, \bfI \big)$ by recursively applying the estimated reverse-time process. Following \textit{classifier-free guidance}~\cite{ho2022classifier}, the implicit update direction $\hat{\bfs}_t$ can be considered in the following form 
\begin{equation}
    \begin{aligned}
        \hat{\bfs}_t  &=   \gamma \bfs_\bftheta(\bfx_t, t, c) + (1 - \gamma )\bfs_\bftheta(\bfx_t, t, \varnothing ),
    \end{aligned}
    \label{eqn:cfg}
\end{equation}
where $\bfs_\bftheta(\bfx_t, t, \varnothing )$ is referred to as as an unconditional model, $\gamma \geq  1$ controls the guidance strength of a condition $c \in \mathcal{C}$. Trivially increasing $\gamma$ will amplify the effect of conditional input.
Condition space $\mathcal{C}$ can be further defined to be text words, called \textit{text prompt}~\cite{nichol2021glide}, fundamentalising current T2I models.

\section{Method}
\label{sec:method}
\begin{figure*}[!h]
    \includegraphics[width=1.\linewidth]{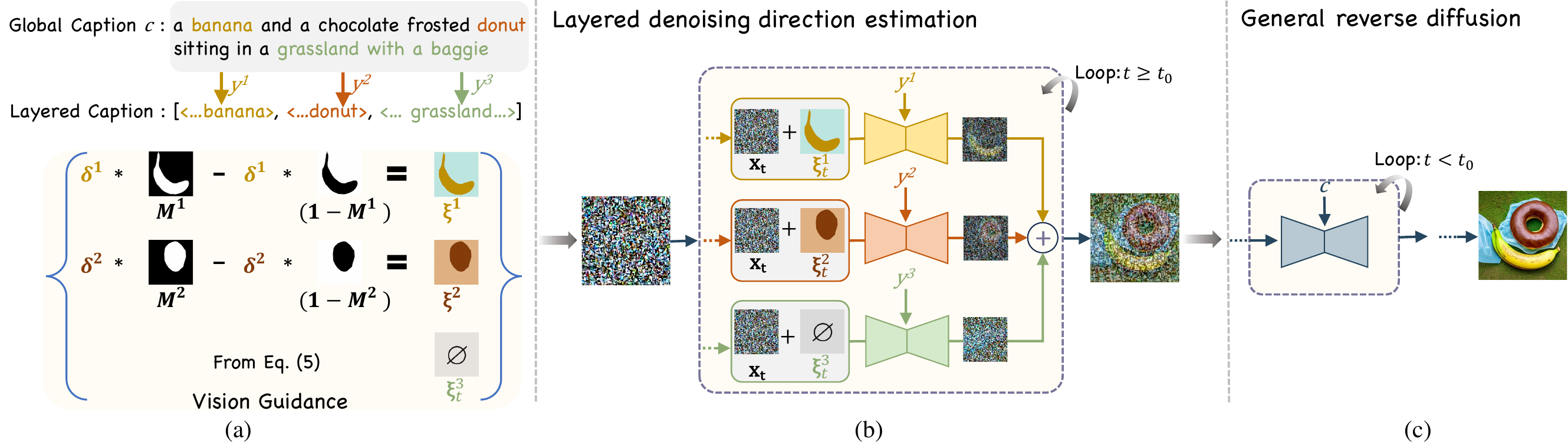}
    \caption{Overview of our framework. (a) For synthesising a sense, the user provides the global caption, the layered caption, as well as the spatial layout entities which are used to construct the vision guidance. LRDiff divides the reverse-time diffusion process into two sections: (b) When $t\geq t_0$, each vision guidance is employed into separate layers to alter the denoising direction, ensuring each object contour generates within specific regions. (c) When $t < t_0$, we perform the general reverse diffusion process to generate texture details that are consistent with the global caption.
    }
    \label{fig:pipline}
    \vspace{-0.25cm}
\end{figure*}

Given a global text caption and the layout condition (bounding boxes or instant masks), our framework can generate images with accurate spatial alignments in a zero-shot paradigm. The remainder of this section is organised as follows: First, we introduce vision guidance, constituting one of the cores of our framework, acting as a foundational spatial cue for the score estimate network to guide the denoising direction of a single visual concept within a specified region. Subsequently, we detail the image-rendering process of LRDiff,  where vision guidance is employed in layers for multiple visual concepts while aligning high-level concepts of the images with the global caption. The overall pipeline is shown in Fig.~\ref{fig:pipline}.

\subsection{Vision Guidance}
\label{vision guidance}
We introduce vision guidance, denoted as $\bm{\xi} \in \mathbb{R}^{h\times w \times D}$, as an additional feature map to the score estimate network, forming $\bfs_\bftheta(\bfx_{(t)}, ~t, ~c, ~\bm{\xi})$. Furthermore, the vision guidance entities are input into the network in a zero-shot form.
A significant advantage of this zero-shot paradigm is that the introduction of the additional condition has no re-training requirement for off-the-shelf conditional diffusion models, thereby substantially reducing computational costs. 
More analysis of vision guidance can be found in the supplementary material.

\vspace{0.15cm}
\noindent \textbf{The Definition.} 
We factorise the vision guidance into two components: a vector $\bm{\delta} \in \mathbb{R}^{D}$ and a binary mask $\bm{\mathcal{M}} \in \{0,1\}^{h \times w}$. Each element $\xi_{j,k,l}$ of $\bm\xi$ is defined as follows:
\begin{equation}
\begin{aligned}
    \xi_{j,k,l} &= \delta_l \cdot \mathcal{M}_{j,k} - \delta_l \cdot (1 - \mathcal{M}_{j,k}) , \\
    &= \delta_l \cdot (2\mathcal{M}_{j,k} - 1),
\end{aligned}
\label{eqn:def_xi}
\end{equation}
where $\mathcal{M}_{j,k}$ is assigned the value 1 if the spatial position $({j,k})$ falls within the expected object region.
For the region containing an object, we add $\bm{\delta}$ to enhance the generation tendency of that object. Conversely, for areas outside the target region, we subtract 
$\bm{\delta}$ to suppress the generation tendency of the object.
The binary mask $\bm{\mathcal{M}}$ can be derived from user input, such as converting a bounding box or instance mask provided by the user into the binary mask. Next, we introduce two distinct approaches to compute the vector $\bm{\delta}$.

\texttt{Constant vector:} A nai\"ve approach for the configuration is to set the vector $\bm{\delta}$ to some constant values. When the diffusion model operates in the RGB space, we can set
$\bm{\delta}$ to constant values corresponding to some colour described by the text prompt (\eg, $[0.3, 0.3, 0.3]$ corresponding to a white colour with transparency). 
When operating in the latent space of VAE, $\bm{\delta}$ can be set to the latent representation of the constant values when operations such as dimension expansion and tensor repeat are required. 
Although the manual adjustment of $\bm{\delta}$ to some constant values is versatile for generating objects with various visual concepts, it necessitates human intervention, such as defining the colour of the object.

\texttt{Dynamic vector:} Beyond simply assigning constant values to $\bm{\delta}$, we propose to dynamically adapt the values of $\bm{\delta}$ based on the input text conditions in order to reduce human intervention during generation. In this context, we consider the implementation of Stable Diffusion~\cite{rombach2022high} wherein text tokens are interconnected with the visual features via cross-attention modules.
At the initial denoising step, \ie, $t=T$, we extract the cross-attention map $\mathbf{A} \in \mathbb{R}^{|c| \times hw}$ from an intermediate layer in the U-Net. For a more straightforward illustration, we will consider the synthesis of an image containing a single object, corresponding to the $i$-th text token from the text prompt $c$. Subsequently, to derive the vector $\bm{\delta}$ in Eq.~\cref{eqn:def_xi}, we perform the following operations:
\begin{equation}
    \begin{aligned}
        \bm{S} &= \{(j,k)| \mathbf{A}_{j,k}^i > \mathrm{Threshold}_K(\mathbf{A}^i)\},\\
        \bm{\delta}  &= \frac{\lambda}{|\bm{S}|}\sum \{ \bfx_t (j,k)| (j,k) \in \bm{S}\},
        \label{eqn:dynamic_xi}
    \end{aligned}
\end{equation}
where $\bfx_t (j,k)$ denotes the element at spatial location $(j,k)$ in $\bfx_t$. The $\sum$ operation sums up all items within the $\bm{S}$ set. Additionally, the operation $\mathrm{Threshold}_K(\cdot)$ selects the $K$-th largest value from the top $K$ values in $\mathbf{A}^i$.
The strength of vision guidance is modulated by the coefficient $\lambda$, alongside the classifier-free guidance coefficient $\gamma$. Given the presence of multiple cross-attention blocks within the score network, we opt to select the block following the down-sampling in each stage and subsequently average their outputs.

\subsection{Layered Rendering}
Considering an upcoming image drawing $n$ objects, we encapsulate all the condition entries of the diffusion model into the following 
\begin{equation}
\begin{aligned}
    \mathrm{Global~Caption:} &~~~~ c, \\
    \mathrm{Layered~Captions:} &~~~~ [y^1(c), \cdots , y^n(c)], \\
    \mathrm{Vision~Guidance: } &~~~~ [\bm\xi^1, \cdots , \bm\xi^n], \\
    \mathrm{Layered~Masks: } &~~~~ [\bm{\mathcal{M}}^1, \cdots , \bm{\mathcal{M}}^n],
\end{aligned}
\end{equation}
where we define a set of mappings $y^i: \mathcal{C} \rightarrow \mathcal{Y}$, and $y^i(c)$~\footnote{We denote $y^i(c)$ by $y^i$ for simplifying notations.} represents the text condition for layer $i$. $\bm\xi^i$ is the vision guidance for the $i_{th}$ object constructed by the $\bm{\mathcal{M}}^i$ and $\delta^i$. Appendix~A from our supplementary material details the implementation of layered caption construction.

As mentioned in Sec.~\ref{intro}, our layered rendering algorithm divides the full reverse-time denoising process into two denoising sections, \ie $[T,\cdots ,t_0], ~[t_0{-}1,\cdots ,1]$.
At each timestep $t \in [T,\cdots ,t_0]$, the denoising process is given by 
\begin{subequations}
\label{eqn:multi_render}
\begin{align}
    \bfx_{t{-}1} &= \tilde{\alpha}_{t} \bfx_{t} + \tilde{\beta}_{t} [\gamma~\bm{\Phi}_t + (1-\gamma)\bfs_\bftheta(\bfx_{t}, t, \varnothing)] + \sigma_{t}\eps_{t}, \label{eqn:multi_render1}\\
    \bm{\Phi}_t &= \sum_{i=1}^n \frac{\bm{\mathcal{M}}^i}{\sum_{i=1}^n \bm{\mathcal{M}}^i} \otimes \bfs_\bftheta(\bfx_{t}{+}\bm{\xi}^i, t, y^i).
    \label{eqn:multi_render2}
\end{align}
\end{subequations}
As shown in Fig.~\ref{fig:pipline}, we construct $\bm{\Phi}_t$ by fusing estimated noises in each layer and sending it into the next denoising loop.
Equation~\ref{eqn:multi_render} assures that the prediction $\bfx_{t{-}1} $ for any step from $[T,\cdots ,t_0]$ is within the data distribution $p_t(\bfx)$, so that the score network $\bfs_\bftheta$ needs no fine-tuning.  
The derivation process of Equation~\ref{eqn:multi_render} is provided in the supplementary material.
We carry out the layered generative process along with vision guidance as expressed in Equation~\ref{eqn:multi_render} for $[T,\cdots ,t_0]$. For timesteps in the second denoising section, \eg, $[t_0{-}1,\cdots ,1]$, we perform the standard denoising process but with the global caption $c$ as the solid condition information, illustrated in the third column in Fig.~\ref{fig:pipline}.
The overall generative process is described by Algorithm~\ref{algo:hlin}. The framework of our method is illustrated in Fig.~\ref{fig:pipline}.  We assign a null value $\varnothing$ to the vision guidance $\bm{\xi}^n$ in the final layer (\eg, the $3^{\mathrm{rd}}$ layer in Fig.~\ref{fig:pipline}), encompassing scene descriptions, such as `a beautiful grassland' or other objects without defined layouts.
\begin{algorithm}
    \caption{\textbf{Layered Rendering Diffusion}}    
    \label{algo:hlin}
    \begin{algorithmic}[1]
    \State \textbf{Input}: $c, ~[y^1, \cdots , y^n], ~[\bm{\mathcal{M}}^1, \cdots , \bm{\mathcal{M}}^n]$,
    \State \quad Pre-trained diffusion model $\bfs_{\bftheta}$; 
    \State \quad Initialise $\bfx_{T}$ \Comment{\textcolor{gray}{Noise initialisation}}
    \State \quad  $[\bm{\xi}^1,...,\bm{\xi}^n]$ $\leftarrow$ Calculate~\cref{eqn:dynamic_xi} 
    \For{$t=T,\cdots,t_0$}
        \State \quad \Comment{\textcolor{gray}{Estimate layered denoising direction}}
        \State \quad  $\bm{\Phi}_t$ $\leftarrow$  Calculate~\cref{eqn:multi_render2} 
        \State \quad $\bfx_{t{-}1}$ $\leftarrow$  Calculate~\cref{eqn:multi_render1}
        \State \quad $\bfx_{t} = \bfx_{t-1}$
    \EndFor
       
    \For{$t=t_0{-}1,\cdots,1$}
           \State \quad \Comment{\textcolor{gray}{Estimate general denoising direction}}
           \State \quad $\hat{\bfs}_t = \bfs_\bftheta(\bfx_t, t) + \gamma \big(\bfs_\bftheta(\bfx_t, t, \mathcal{C}) - \bfs_\bftheta(\bfx_t, t) \big)$
         \State \quad $\bfx_{t-1} = \tilde{\alpha}_t \bfx_{t} + \tilde{\beta}_t \hat{\bfs}_t + \sigma_t \eps_t$
         \State \quad $\bfx_{t} = \bfx_{t-1}$
    \EndFor
    \State \textbf{Output}: $\bfx_{0}$
    \end{algorithmic}
    \label{algo:optimization}
\end{algorithm}



\section{Experiments}

\noindent \textbf{Dataset and Implementation Details}. All the experiments are run on a single NVIDIA Tesla V100. Unless specified otherwise, we use the DDIM sampler~\cite{song2020denoising} with 50 sampling steps for the reverse diffusion process with a fixed guidance scale of 7.5; $t_0$ is set to 15 by default. We construct our dataset by selecting 1134 captions with one or more objects and corresponding bounding boxes or instance masks from the MS-COCO validation set~\cite{lin2014microsoft}. For a fair comparison, we implement LRDiff based on a diffusion model with a version similar to that used by other methods, aiming to eliminate differences in results caused by variations in the capabilities of the foundational diffusion model.


\vspace{0.15cm}
\noindent \textbf{Evaluation Metrics.} 
For evaluating synthesised images with both bounding box and instance mask inputs, we employ two distinct metrics: image-score and align-score.
The image-score specifically measures the fidelity of the synthesised image to the text prompt, incorporating sub-indicators such as T2I-Sim~\cite{xie2023boxdiff} and the CLIP score~\cite{hessel2021clipscore} for a nuanced assessment.
On the other hand, the align-score evaluates the image's spatial alignment with the given layout condition, using the AP results predicted by YOLOv4 as a benchmark for alignment accuracy. Additionally, regarding instance mask inputs, we assess the precision of object contours using the IoU scores produced by employing YOLOv7~\cite{wang2023yolov7} and the ground truth.

\begin{table}[!h]
  \centering
  \begin{minipage}[b]{0.477\linewidth}
    \centering
    \scriptsize
    \renewcommand{\arraystretch}{1.1}
    \addtolength{\tabcolsep}{-1pt}
    \begin{tabular}{lccccc}
    \toprule
    \multicolumn{1}{c}{\multirow{2}{*}{Bounding Box}} & \multicolumn{2}{c}{Image-Score} & \multicolumn{3}{c}{Align-Score} \\ \cmidrule(r){2-3} \cmidrule(r){4-6}
    \multicolumn{1}{c}{}    & T2I-Sim$\uparrow$   & CLIP$\uparrow$   & $\mathrm{mAP}$$\uparrow$&$\mathrm{AP}_{50}$$\uparrow$ & $\mathrm{AP}_{75}$$\uparrow$\\ \cmidrule(r){1-3} \cmidrule(r){4-6}
    SD~\cite{rombach2022high}   & 0.292     & 0.316    & -    & -  &  - \\
TwFA~\cite{yang2022modeling}          & 0.210     & 0.179   & \cellcolor{gray!20}9.9   & \cellcolor{gray!20}16.3 &\cellcolor{gray!20}9.0 \\
eDiffi-Pww~\cite{balaji2022ediffi}       & 0.279  & 0.299  & 4.2    & 8.6 & 4.0 \\
BoxDiff~\cite{xie2023boxdiff}     & 0.295   & 0.319    & 5.8& 17.2 & 3.0 \\
\cellcolor{yellow!20}LRDiff(Ours)  &\cellcolor{yellow!20} 0.281 & \cellcolor{yellow!20}0.292  & \cellcolor{yellow!20}\textbf{17.4}   &\cellcolor{yellow!20}\textbf{35.6} & \cellcolor{yellow!20}\textbf{15.5} \\
\bottomrule
\end{tabular}
\vspace{0.15cm}
\caption{The quantitative results of bounding box input.}
\label{box_results_quantitative}
  \end{minipage}
  \vspace{-0.15cm}
  \hfill
  \begin{minipage}[b]{0.477\linewidth}
    \centering
    \scriptsize
    \renewcommand{\arraystretch}{1.1}
    \addtolength{\tabcolsep}{-1pt}
    \begin{tabular}{lccccc}
    \toprule
    \multicolumn{1}{c}{\multirow{2}{*}{Instance Mask}} & \multicolumn{2}{c}{Image-Score} & \multicolumn{3}{c}{Align-Score} \\ \cmidrule(r){2-3} \cmidrule(r){4-6}
    \multicolumn{1}{c}{}    & T2I-Sim$\uparrow$   & CLIP$\uparrow$   & $\mathrm{AP}_{50}$$\uparrow$ & $\mathrm{AP}_{75}$$\uparrow$   & IOU$\uparrow$\\ \cmidrule(r){1-3} \cmidrule(r){4-6}
    SD~\cite{rombach2022high}       & 0.292          & 0.316   & -     & -  & - \\
eDiffi-Pww~\cite{balaji2022ediffi}     & 0.287    & 0.304   & 16.5 &  6.4 &33.11\\
MultiDiff~\cite{bar2023multidiffusion}     & 0.277   &0.281   & 30.9    & 8.2 &47.59\\
DenseDiff.~\cite{kim2023dense}   & 0.289  &  0.310 & 11.3 & 1.3 & 27.65\\
\cellcolor{yellow!20}LRDiff(Ours)  & \cellcolor{yellow!20}0.280  & \cellcolor{yellow!20}0.295   &\cellcolor{yellow!20} \textbf{35.0}     & \cellcolor{yellow!20} \textbf{15.4}&\cellcolor{yellow!20}\textbf{49.06} \\
\bottomrule
\end{tabular}
\vspace{0.15cm}
\caption{The
quantitative results of instance mask input.}
\label{mask_results_quantitative}
  \end{minipage}
  \vspace{-0.15cm}
\end{table}

\vspace{-1cm}
\subsection{Bounding boxes as layout condition}
\label{box comparison}
\noindent \textbf{Quantitative comparison.} In Table~\ref{box_results_quantitative}, we present a comparative analysis of our method against BoxDiff~\cite{xie2023boxdiff}, eDiffi-Pww~\cite{balaji2022ediffi}, and TwFA~\cite{yang2022modeling}. Our evaluation revealed a trade-off relationship between the align-score and image-score among the compared methods. For instance, while TwFA~\cite{yang2022modeling} achieved superior alignment scores compared to other diffusion methods, its limited image generation capabilities led to lower-quality generated backgrounds. Consequently, this facilitated easier foreground differentiation by YOLO~\cite{bochkovskiy2020yolov4}, resulting in higher detection metrics. To ensure fairness in comparisons, we established SD~\cite{rombach2022high} as the baseline for the image-score. Notably, our results exhibited higher AP values in contrast to eDiffi-Pww and TwFA, while closely aligning with the baseline in terms of image-score. Furthermore, our image-score values are close to those of BoxDiff, but our three alignment score sub-metrics exceed BoxDiff by 11.6\%, 17.6\%, and 12.5\%, respectively.

\begin{figure*}[!h]
    \centering
    \includegraphics[width=1.0\linewidth]{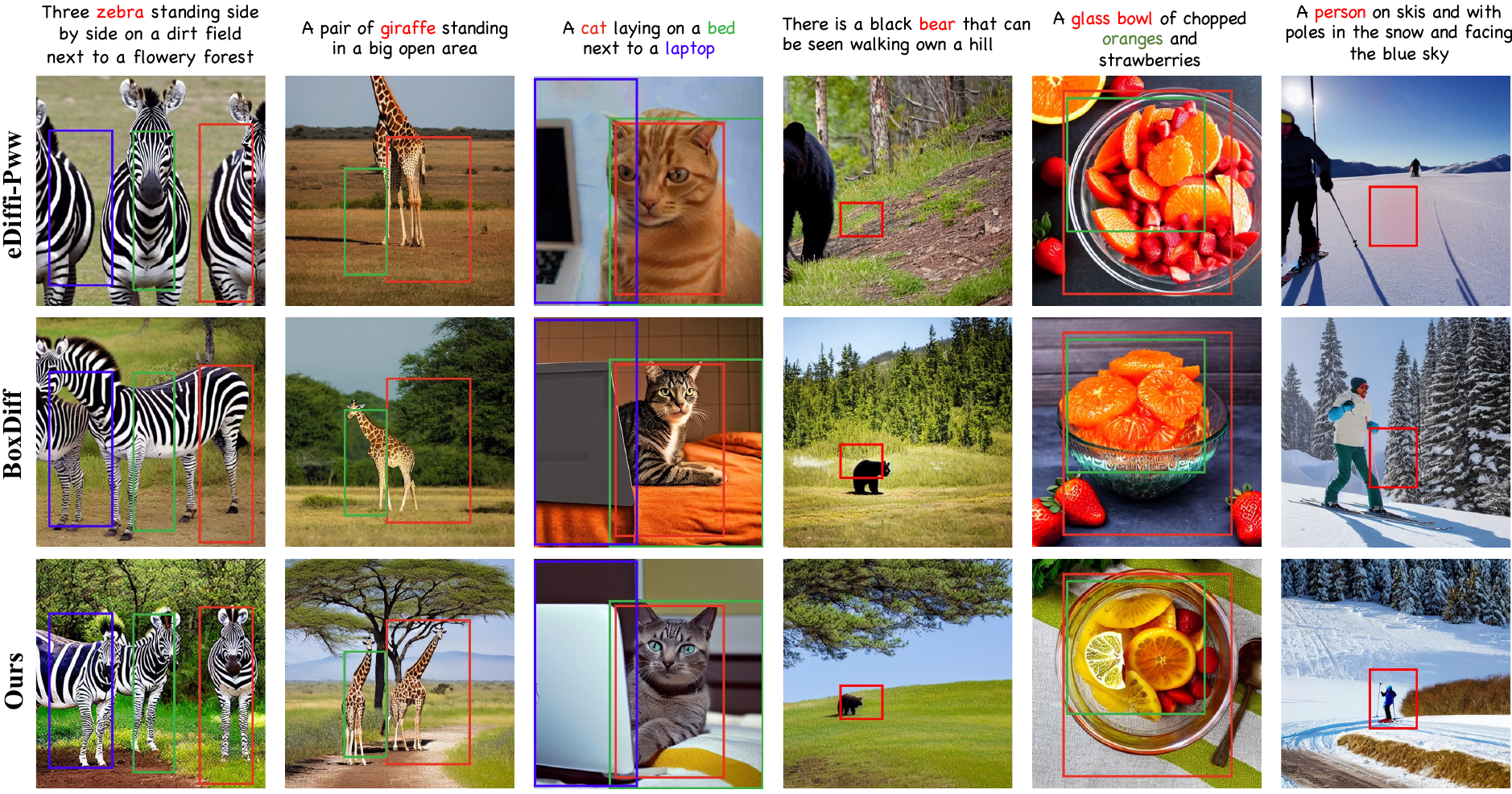}
    \caption{Qualitative comparisons of methods that use bounding box entities as the spatial condition. Our results show better spatial alignments than other methods. 
    }
    \label{box_results_qualitative}
\end{figure*}

\vspace{0.15cm}
\noindent \textbf{Qualitative comparison.} Fig.~\ref{box_results_qualitative} presents qualitative comparisons among various methods in a multi-object layout. 
According to the results, LRDiff can effectively mitigate the issue of visual blending between adjacent objects within the same category, which is a challenge for other methods such as BoxDiff~\cite{xie2023boxdiff} and eDiffi-Pww~\cite{balaji2022ediffi}. This effectiveness is exemplified in the synthesis of giraffes, as shown in the second column of Fig.~\ref{box_results_qualitative}. Furthermore, our method surpasses eDiffi-Pww when synthesising images within intricate layouts. Notably, in the third column, where a cat, bed and laptop are closely arranged, our LRDiff proficiently handles mutual occlusion, underscoring its capability to manage complex scene compositions. Additionally, our approach shows high fidelity in generating small-scale objects, which remains challenging for the listed methods that rely on manipulating cross-attention maps to control layout.

\subsection{Instance masks as layout condition}
\label{quantitaive comparsion}

 \noindent \textbf{Quantitative comparison.} The instance masks, serving as layout guides, provide both positional and contour details. As highlighted in Table~\ref{mask_results_quantitative}, our approach outperforms other diffusion-based zero-shot techniques \cite{kim2023dense,bar2023multidiffusion,balaji2022ediffi} in the IoU metric. Similar to the analysis in subsection.~\ref{box comparison}, there is a trade-off correlation between the image-score and align-score. Our results significantly outperform eDiffi-Pww and DenseDiff by over 16\% and 22\%, respectively, in the IOU metric while closely aligning with their image-score outcomes.
 The observed low image-score in MultiDiff \cite{bar2023multidiffusion} is attributed to its limited interaction with global captions, leading to a `copy-paste-like' phenomenon. In contrast, our method integrates all layers and interacts with global captions after delineating object outlines in layers. Consequently, compared to MultiDiff \cite{bar2023multidiffusion}, our method achieves a closer alignment with the baseline image-score, surpassing it by 1.6\% in the mIoU metric. Furthermore, substantial enhancements in both $AP_{50}$ and $AP_{75}$ indicators by over 9\% each signify the accurate alignment and positioning of generated objects within the specified mask area, validating the efficacy of our approach.

\begin{figure*}[!h]
    \centering
    \includegraphics[width=1.0\linewidth]{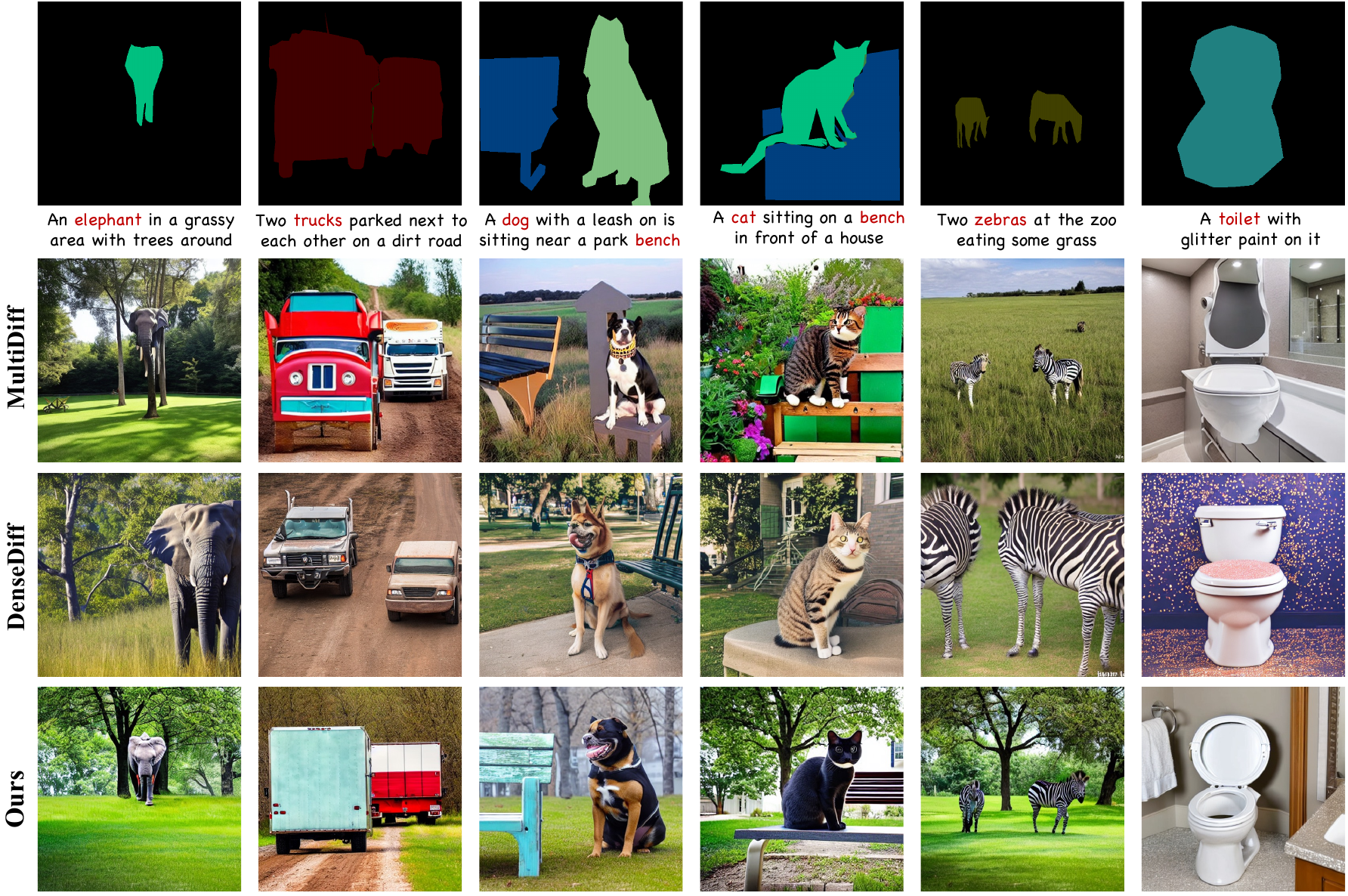}
    \caption{Qualitative comparisons of methods that use instance mask entities as the spatial condition. Our results show better spatial alignments than other methods. 
    }
    \label{mask_results_qualitative}
\end{figure*}

\vspace{0.15cm}
\noindent \textbf{Qualitative comparison.} Fig.\ref{mask_results_qualitative} presents qualitative comparisons between our method and others across both single and multi-object layouts. The results highlight our effectiveness in generating small-scale objects, which is still a challenge for DenseDiff \cite{kim2023dense}.
Examples to justify our effectiveness are provided in the first and fifth columns of  Fig.\ref{mask_results_qualitative} where the synthesised elephants and zebras are faithful to the shapes provided in the layout entity. 
MultiDiff~\cite{bar2023multidiffusion} demonstrates high effectiveness in achieving precise spatial alignment within images. However, it exhibits limitations in harmonising the overall image composition, occasionally resulting in a `copy-paste' result. This inadequacy is particularly evident in the first and last columns of the figure, where an elephant appears inserted into a tree and a toilet lacks seamless integration with the surrounding environment. Conversely, our method ensures precise image layout accuracy by employing visual guidance and achieves seamless integration throughout the entire image due to the fusion of all layers and interaction with the global captions.

\vspace{-0.45cm}
\subsection{Ablation Study}
\label{ablation study}
In this section, we conduct experiments primarily to demonstrate the necessity of our designed vision guidance and the different impacts on different ways of calculating $\delta_l$. Additionally, we showcase the impacts of our method across diverse $t_0$ values. Further ablation experiments and more results are provided in the supplementary material.

\begin{figure*}[!h]
    \centering
    \includegraphics[width = 1.0\linewidth]{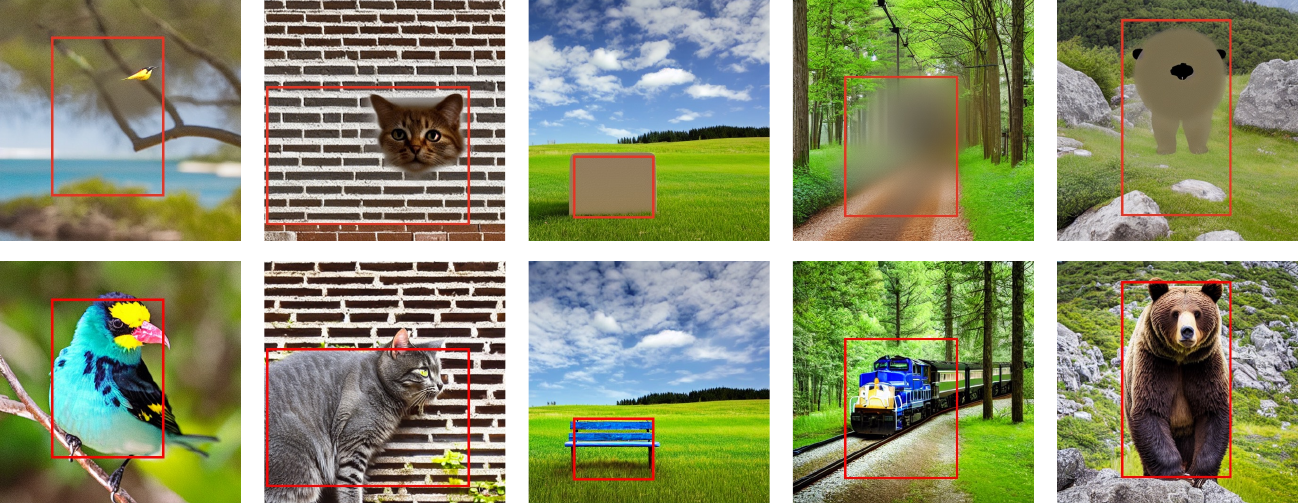}
    \caption{The bounding boxes condition. The first row shows the results using setting~\#1 and the second row shows the results using setting~\#2.}
    \label{with vs without guidance_box}
    \vspace{-1cm}
\end{figure*}
\begin{figure*}[!h]
    \centering
    \includegraphics[width = 1.0\linewidth]{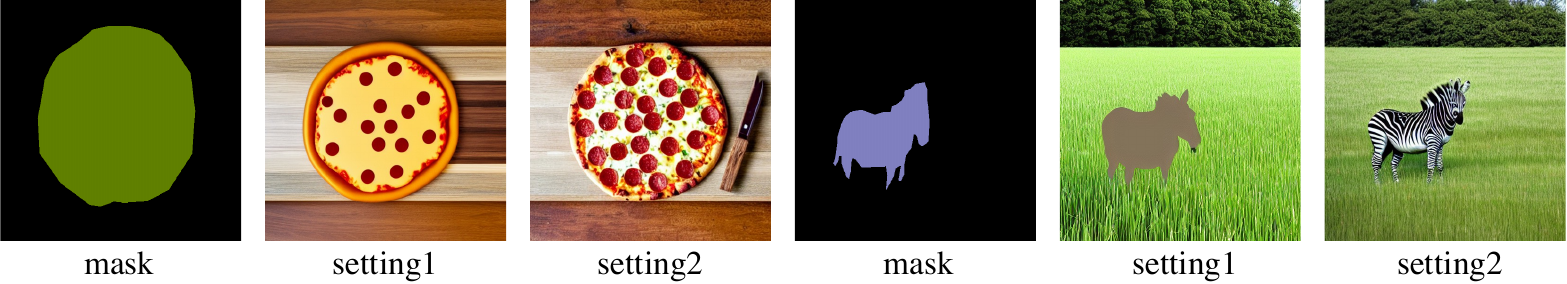}
    \caption{The instant mask condition. The different results of using setting~\#1 and setting~\#2.}
    \label{with vs without guidance_mask}
\end{figure*}
\begin{figure}[!h]
    \centering
    \includegraphics[width=0.8\linewidth]{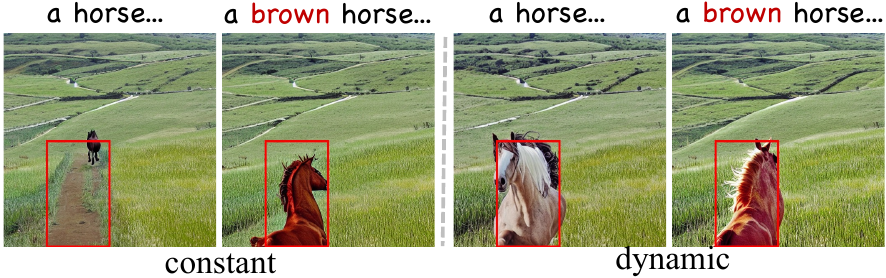}
    \caption{The dynamic vector strategy shows high spatial alignment to the given layout, eliminating the need for human intervention .}
    \label{fig:ablation-study}
    \vspace{-0.35cm}
\end{figure}

\vspace{0.15cm}
\noindent \textbf{The necessity of vision guidance.} 
To investigate the necessity of visual guidance, we conduct an experiment by removing the visual guidance within the target area and only suppressing the vision guidance outside the target area. Correspondingly, we modify Eq.~\ref{eqn:def_xi} as follows:
\begin{equation}
\begin{aligned}
\xi_{j,k,l} &= - \delta_l \cdot (1 - \mathcal{M}_{j,k}).
\end{aligned}
\label{eqn:def_xi_background}
\end{equation}
To write concisely, we name the two ways of constructing $\bm\xi$ using Eq.~\ref{eqn:def_xi_background} and Eq.~\ref{eqn:def_xi} respectively as setting \#1 and setting \#2. 
\begin{table}[!h]
  \centering
  \begin{minipage}[b]{0.477\linewidth}
    \centering
    \scriptsize
    \renewcommand{\arraystretch}{1.1}
    \addtolength{\tabcolsep}{-1pt}
    \begin{tabular}{lccccc}
    \toprule
    \multicolumn{1}{c}{\multirow{2}{*}{Bounding Box}} & \multicolumn{2}{c}{Image-Score} & \multicolumn{3}{c}{Align-Score} \\ \cmidrule(r){2-3} \cmidrule(r){4-6}
    \multicolumn{1}{c}{}    & T2I-Sim$\uparrow$   & CLIP$\uparrow$   & $\mathrm{mAP}$$\uparrow$&$\mathrm{AP}_{50}$$\uparrow$ & $\mathrm{AP}_{75}$$\uparrow$\\ \cmidrule(r){1-3} \cmidrule(r){4-6}
    setting \#1 & 0.1967& 0.1642&1.3 &3.8 & 0.7\\
    \cellcolor{gray!20}{setting \#2} &\cellcolor{gray!20}{0.281}&\cellcolor{gray!20}{0.295} & \cellcolor{gray!20}{17.4}&\cellcolor{gray!20}{35.6} &\cellcolor{gray!20}{15.5}\\
    \bottomrule
    \end{tabular}
    \vspace{0.15cm}
    \caption{These results depict the outcome when vision guidance is not applied within the target areas under the bounding box condition.}
    \label{noenhance_box}
  \end{minipage}
  \vspace{-0.25cm}
  \hfill
  \begin{minipage}[b]{0.477\linewidth}
    \centering
    \scriptsize
    \renewcommand{\arraystretch}{1.1}
    \addtolength{\tabcolsep}{-1pt}
    \begin{tabular}{lccccc}
    \toprule
    \multicolumn{1}{c}{\multirow{2}{*}{Instance Mask}} & \multicolumn{2}{c}{Image-Score} & \multicolumn{3}{c}{Align-Score} \\ \cmidrule(r){2-3} \cmidrule(r){4-6}
    \multicolumn{1}{c}{}    & T2I-Sim$\uparrow$   & CLIP$\uparrow$   & $\mathrm{AP}_{50}$$\uparrow$ & $\mathrm{AP}_{75}$$\uparrow$   & IOU$\uparrow$\\ \cmidrule(r){1-3} \cmidrule(r){4-6}
    setting \#1& 0.2599 & 0.2767& 5.2& 0.8&28.16 \\
    \cellcolor{gray!20}{setting \#2}&\cellcolor{gray!20}{0.280}&\cellcolor{gray!20}{0.295} & \cellcolor{gray!20}{35.0}& \cellcolor{gray!20}{15.4}&\cellcolor{gray!20}{49.06}\\
    \bottomrule
    \end{tabular}
    \vspace{0.15cm}
    \caption{These results depict the outcome when vision guidance is not applied within the target areas under the instance mask condition. }
    \label{noenhance_mask}
  \end{minipage}
  \vspace{-0.25cm}
\end{table}
First, the qualitative difference between the two ways can be found in Fig.~\ref{with vs without guidance_box} and Fig.~\ref{with vs without guidance_mask}. The results show that suppressing visual guidance outside the target region can effectively eliminate the tendency to generate objects. However, this does not relatively enhance the tendency to generate objects within the target region, as the features in the unmodified region may inherently lack the capability to generate objects. Interestingly, the use of setting \#2, as demonstrated in Fig.~\ref{with vs without guidance_mask}, has a certain effectiveness on simple objects, such as pizza, where realism is lacking. This is attributed to the mask providing certain additional shape information. Furthermore, Tables \ref{noenhance_box} and \ref{noenhance_mask} present a direct comparison between the results obtained with and without vision guidance in the target area. The outcomes strongly indicate that the absence of such guidance renders the results nearly unusable. Considering the findings from these tables and figures, it can be concluded that vision guidance significantly impacts the effectiveness of the generated outcomes.
\noindent \textbf{Constant vs. Dynamic.} 
We visually compared two approaches for calculating the vector $\bm{\delta}$ in vision guidance (Fig.~\ref{fig:ablation-study}). In constant mode, if the prompt does not specify colour (the first column), $\bm{\delta}$ defaults to brown, which is a common colour for horses. However, this may mislead, as brown becomes associated with roads instead of horses. The reason is that `brown information' is not effectively associated with the concept of a horse without explicit cue. The constant mode excels when colour is explicitly mentioned (second column). On the contrary, the dynamic vector strategy consistently aligns with the layout, making it preferable for real-world applications, and eliminating the need for human intervention in specifying object colours.

\begin{figure*}[!h]
    \centering
    \includegraphics[width=1.\linewidth]{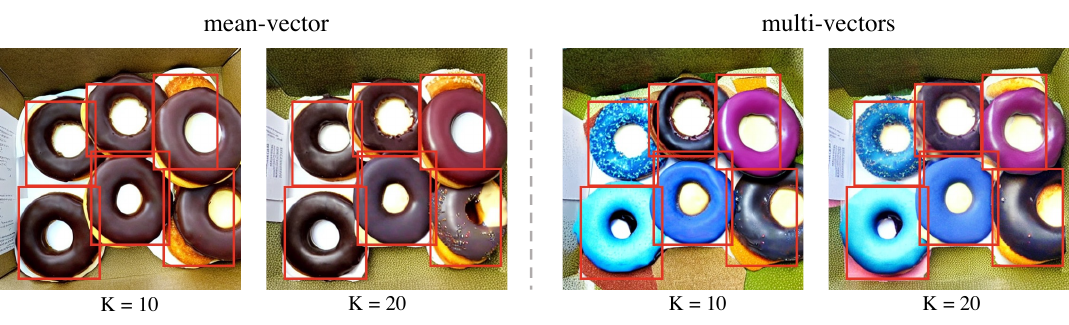}
    \caption{The varying impacts of employing different construction methods. The generated images with `random-vectors' are more diverse in terms of textures and patterns for different objects.}
    \label{scatter vs fusing}
    \vspace{-0.15cm}
\end{figure*}

\noindent \textbf{Constructing $\delta$ with $K$ random vectors.} From the above experiments, we observe that dynamic vision guidance serves as an effective cue for providing layout information. Furthermore, we present an alternative implementation of $\delta$ in dynamic mode, utilising $K$ random vectors. Specifically, for each position in the object area, we fill the position with a vector randomly sampled from set $\mathbf{S}$ from~Eq.~\ref{eqn:dynamic_xi}. We denote this as `random-vectors', distinguishing it from the `mean-vector' method mentioned in Sec.~\ref{vision guidance}.
\begin{table}[!h]
  \centering
  \begin{minipage}[b]{0.47\linewidth}
    \centering
    \scriptsize
    \renewcommand{\arraystretch}{1.1}
    \addtolength{\tabcolsep}{-0.5pt}
    \begin{tabular}{lccccc}
    \toprule
     \multicolumn{1}{c}{\multirow{2}{*}{Bounding Box}}  & \multicolumn{2}{c}{Image-Score} & \multicolumn{3}{c}{Align-Score} \\ \cmidrule(r){2-3} \cmidrule(r){4-6}
     & T2I-Sim$\uparrow$   & CLIP$\uparrow$   & $\mathrm{mAP}$$\uparrow$&$\mathrm{AP}_{50}$$\uparrow$ & $\mathrm{AP}_{75}$$\uparrow$\\ \cmidrule(r){1-3} \cmidrule(r){4-6}
     $\text{random-vectors}_{10}$& 0.278& 0.293&17.0 &30.0 &13.9 \\
     $\text{mean-vector}_{10}$& 0.284& 0.298& 17.2& 33.2&14.6 \\
     $\text{random-vectors}_{20}$&0.272 & 0.287& 17.1& 32.9& 14.5\\
     $\cellcolor{gray!20}{\text{mean-vector}_{20}}$& \cellcolor{gray!20}{0.281}&\cellcolor{gray!20}{0.295} & \cellcolor{gray!20}{17.4}&\cellcolor{gray!20}{35.6} &\cellcolor{gray!20}{15.5} \\
    \bottomrule
    \end{tabular}
    \vspace{0.15cm}
    \caption{Comparison of `random-vectors' and `mean-vector' with bounding box entities. The \textcolor{gray}{grey} line indicates the results in Table~\ref{box_results_quantitative}. The subscript denotes the choice of value $K$.}
    \label{dynamic_vectors_box}
  \end{minipage}
\vspace{-0.25cm}
  \hfill
  \begin{minipage}[b]{0.47\linewidth}
    \centering
    \scriptsize
    \renewcommand{\arraystretch}{1.1}
    \addtolength{\tabcolsep}{-0.5pt}
    \begin{tabular}{lccccc}
    \toprule
    \multicolumn{1}{c}{\multirow{2}{*}{Instance Mask}}  & \multicolumn{2}{c}{Image-Score} & \multicolumn{3}{c}{Align-Score} \\ \cmidrule(r){2-3} \cmidrule(r){4-6}
       & T2I-Sim$\uparrow$   & CLIP$\uparrow$   & $\mathrm{AP}_{50}$$\uparrow$ & $\mathrm{AP}_{75}$$\uparrow$   & IOU$\uparrow$\\ \cmidrule(r){1-3} \cmidrule(r){4-6}
     $\text{random-vectors}_{10}$& 0.281& 0.296& 36.5&19.2 &49.02 \\
     $\text{mean-vector}_{10}$&0.277 &0.287 &37.9 &19.8 & 49.41\\
     $\text{random-vector}_{20}$&0.278 &0.292 & 40.5& 21.0 &49.54 \\
     $\cellcolor{gray!20}{\text{mean-vector}_{20}}$& \cellcolor{gray!20}{0.280}&\cellcolor{gray!20}{0.295} & \cellcolor{gray!20}{35.0}& \cellcolor{gray!20}{15.4}&\cellcolor{gray!20}{49.06} \\
    \bottomrule
    \end{tabular}
    \vspace{0.15cm}
    \caption{Comparison of `random-vectors' and `mean-vector' with instance mask entities. The \textcolor{gray}{grey} line shows the results in Table~\ref{mask_results_quantitative}. The subscript denotes the choice of value $K$.}
    \label{dynamic_vectors_mask}
  \end{minipage}
  \vspace{-0.25cm}
\end{table}
Table~\ref{dynamic_vectors_box} and Table~\ref{dynamic_vectors_mask} highlight the remarkable similarity in results obtained through these two different approaches to constructing $\delta$. The `mean-vector' method better aligns with the input text compared to the `random-vectors'. The results also show that $K$ is set to 10, the image-score outperforms that achieved when $K$ is set to 20. However, to ensure a balanced performance between image-score and align-score, we opt for the `mean-vector' method with $K$ set to 20. The `random-vectors' approach exhibits advantages in achieving texture diversity among multiple objects of the same category within a scene. Illustrated in Fig.~\ref{scatter vs fusing}, an example clarifies the divergent visual effects of employing these two approaches. 
However, in most cases, the `mean-vector' method suffices entirely.

\begin{figure}[!t]
    \centering
    \includegraphics[width=0.8\linewidth]{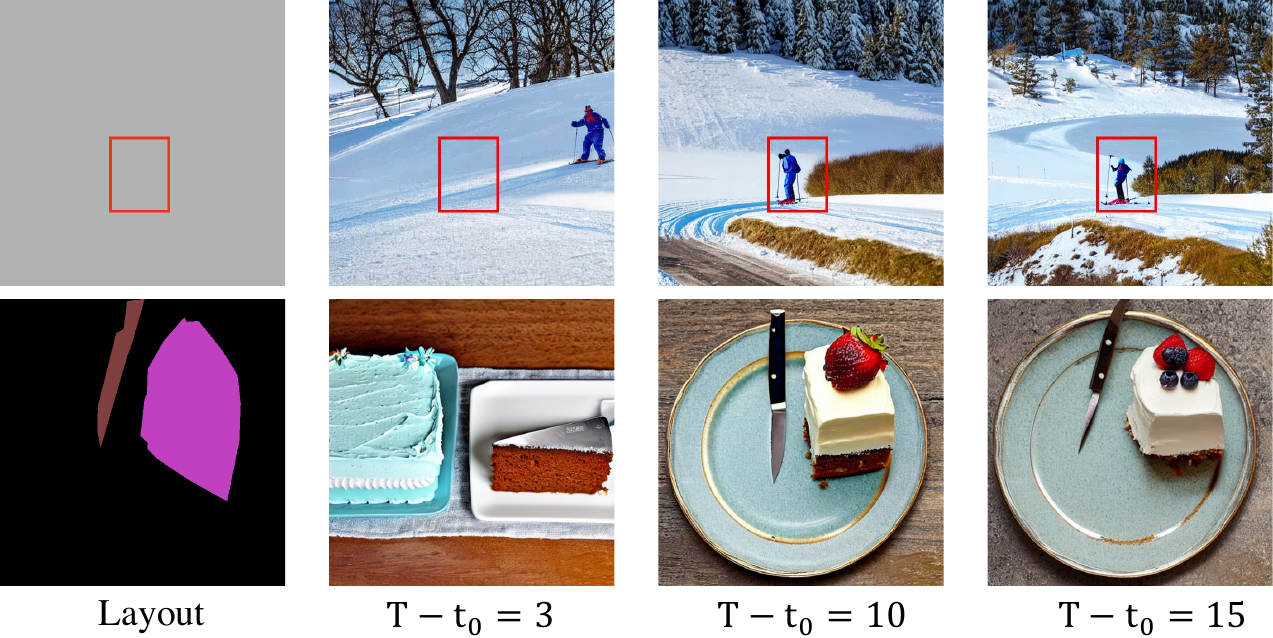}
    \caption{The impacts of different $t_0$. A short first denoising period can result in less precise spatial alignment}
    \label{fig:ablation-study2}
    \vspace{-0.8cm}
\end{figure}

\vspace{0.15cm}
\noindent \textbf{The impact of different $t_0$.}  LRDiff divides the reverse diffusion process into two denoising sections: $[T,\cdots ,t_0], ~[t_0{-}1,\cdots ,1]$. As we analyse in Fig.~\ref{fig:ablation-study2}, a longer first denoising period (\ie, $|T - t_0|$) will generate images of higher spatial aliment with the given spatial layout. However, a short first denoising period can result in less precise spatial alignment. We attribute this raised deviation to the ambiguity in object shapes during the early denoising steps, where they are particularly susceptible to noise from other layers. We observe that setting $|T - t_0| = 15$ usually gives some decent results. The lower part in Fig.~\ref{fig:ablation-study2} shows that when setting $|T - t_0| = 15$ the knife's shape fits better than the others. 
\section{Other applications}
\begin{figure}[!t]
\centering
\includegraphics[width=0.8\linewidth]{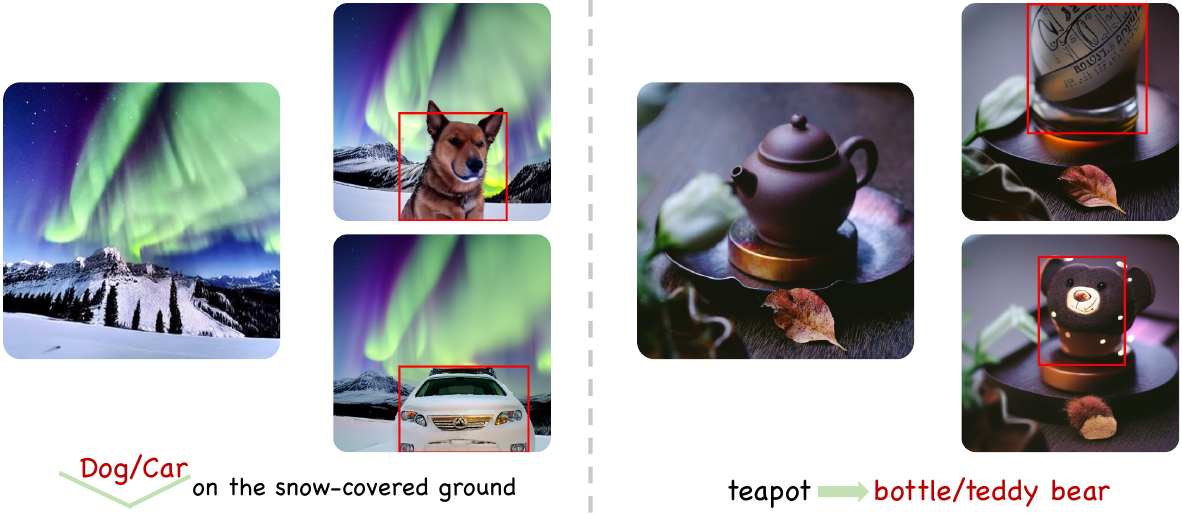}
    \caption{Example results of controlled image editing. Insert or replace an object at a specific location.}
    \label{otherapplication}
\end{figure}

We extend our framework to controllable image editing. We employ the DDIM inversion technique~\cite{song2020denoising} to obtain the noise latent representation. This inverted latent noise serves as the final layer (\eg, the $3^{\mathrm{rd}}$ layer in Fig.~\ref{fig:pipline}) in our pipeline, enabling further processing. As demonstrated in Fig.\ref{otherapplication},
our method allows inserting or replacing objects of various sizes at different locations within a provided image. Thanks to our layered rendering technology, the edited images seamlessly fit the given prompts, while keeping other areas of content as unchanged as possible.
\section{Dissuasion and limitations}
While the proposed framework has delivered refined controllability for layout-to-image tasks, certain limitations exist that necessitate further exploration. First, the spatial conditional inputs are limited to bounding boxes and object masks. We could extend our framework to include other layout conditions, such as key points refining human poses and depth maps providing 3D information about the scene. Additionally, we lack discussion of the sensitivity of the $t_0$ to different types of scenes, such as indoor and outdoor scenes. In our future work, we may address this by using learnable $t_0$ by optimising reward model~\cite{fan2024reinforcement} to enhance the robustness of our method across various scenes. In our experiments, we construct our framework with SD v1.5. Moving forward, we can explore our method with other versions of diffusion models with enhanced capabilities~\cite{podell2023sdxl} or higher speed~\cite{sauer2023adversarial}. Another promising direction involves customised generation under additional layout conditions by combining with a personalisation method, such as DreamBooth~\cite{ruiz2023dreambooth}.

\section{Conclusion}
In this paper, we introduce a universal framework designed to generate results aligned with the input spatial layout conditions while avoiding the blending of multiple visual concepts. The proposed framework, termed LRDiff, comprises a two-stage Layered Rendering Diffusion, establishing an image-rendering process with multiple layers. It incorporates an innovative concept known as vision guidance, playing a crucial role in achieving precise spatial alignment for individual objects in a zero-shot paradigm. The effectiveness of our method has been demonstrated through the experiments, highlighting the superior capabilities of our framework. Additionally, we explored various approaches to constructing vision guidance. Our technology finds applications in three domains: bounding box-to-image, instance mask-to-image, and controllable image editing. We also extended our framework to controllable image editing.


\clearpage  

%
%
\bibliographystyle{splncs04}
\bibliography{main}

\newpage
\appendix

\section{The Derivation of Layered Rendering}
In order to ensure that the prediction $\tilde{\bfx}_{t{-}1}$ for any step from $[T,\cdots ,t_0]$ is within the data distribution $p_t(\bfx)$ with the pre-trained score network $\bfs_\bftheta$, the Layered Rendering method (\ie, Eq.(7) from the main paper) is proposed by solving the following objective
\begin{equation}
     \arg\min_{\tilde{\bfx}_{t} \in \mathcal{I}_{t}} \sum_{i=1}^n \left \| \bm{\mathcal{M}}_i \otimes \left ( \tilde{\bfx}_{t{-}1} - \bfx_{t{-}1}^i \right ) \right \|^2,
\end{equation}
where $\{\bfx_{t{-}1}^i\}_{i=1}^n$ is referred to as the results of $n$ independent denoising processes at timestep $t$; $\tilde{\bfx}_{t{-}1}$ is referred to as the desired result that integrates information from $\{\bfx_{t{-}1}^i\}_{i=1}^n$ based on the provided region contains from $\{\bm{\mathcal{M}}_i\}_{i=1}^n$.

\noindent The analytical solution to this objective is given by
\begin{align}
   \sum_{i=1}^n  \bm{\mathcal{M}}_i \otimes \tilde{\bfx}_{t{-}1} &= \sum_{i=1}^n \bm{\mathcal{M}}_i \otimes \bfx_{t{-}1}^i
\end{align}
\begin{equation}
       \therefore  \tilde{\bfx}_{t{-}1} = \frac{1}{\sum_{i=1}^n \bm{\mathcal{M}}_i} \sum_{i=1}^n \bm{\mathcal{M}}_i \otimes \bfx_{t{-}1}^i = \sum_{i=1}^n \frac{\bm{\mathcal{M}}^i}{\sum_{i=1}^n \bm{\mathcal{M}}^i} \otimes \bfx_{t{-}1}^i .
       \label{eqn: anal}
\end{equation}

\noindent According to the reverse-time diffusion definition,
\begin{equation}
    \bfx_{t{-}1}^i   \approx  \tilde{\alpha}_t \bfx_{t} + \underbrace{\tilde{\beta}_t \hat{\bfs}_t^i}_{\text{estimated direction pointing to } \bfx_t^i} + \underbrace{\sigma_t \eps_t^i}_{\text{random noise}},
\end{equation}
we have
\begin{equation}
\begin{aligned}
       \tilde{\bfx}_{t{-}1} &= \sum_{i=1}^n \frac{\bm{\mathcal{M}}^i}{\sum_{i=1}^n \bm{\mathcal{M}}^i} \otimes \left (\tilde{\alpha_t} \bfx_t^i + \tilde{\beta_t}\hat{\bfs}_t^i +\sigma_t\epsilon_t^i \right )\\
       &= \tilde{\alpha_t}\left(\sum_{i=1}^n \frac{\bm{\mathcal{M}}^i}{\sum_{i=1}^n \bm{\mathcal{M}}^i} \otimes \bfx_t^i \right) + \tilde{\beta_t} \left(\sum_{i=1}^n \frac{\bm{\mathcal{M}}^i}{\sum_{i=1}^n \bm{\mathcal{M}}^i} \otimes \hat{\bfs}_t^i  \right) + \sigma_t \left(\sum_{i=1}^n \frac{\bm{\mathcal{M}}^i}{\sum_{i=1}^n \bm{\mathcal{M}}^i} \otimes  \eps_{t}^i \right) \\
       &= \tilde{\alpha_t} \tilde{\bfx}_{t} + \tilde{\beta_t}\left(\sum_{i=1}^n \frac{\bm{\mathcal{M}}^i}{\sum_{i=1}^n \bm{\mathcal{M}}^i} \otimes \hat{\bfs}_t^i  \right) + \sigma_t \tilde{\epsilon}_t \quad \quad (\mathrm{Apply}~\cref{eqn: anal} )\\
     &= \tilde{\alpha}_{t} \tilde{\bfx}_{t} + \tilde{\beta}_{t} \left\{\sum_{i=1}^n \frac{\bm{\mathcal{M}}^i}{\sum_{i=1}^n \bm{\mathcal{M}}^i} \otimes \left[\gamma~\bfs_\bftheta(\tilde{\bfx}_t{+}\bm{\xi}^i, t, y^i) + (1-\gamma)\bfs_\bftheta(\tilde{\bfx}_{t}, t, \varnothing)\right] \right\} + \sigma_{t}\tilde{\eps}_{t}  \quad\quad\quad (\mathrm{Apply~CFG})\\
     &= \tilde{\alpha}_{t} \tilde{\bfx}_{t} + \tilde{\beta}_{t} \left\{\gamma\left[ \sum_{i=1}^n \frac{\bm{\mathcal{M}}^i}{\sum_{i=1}^n \bm{\mathcal{M}}^i} \otimes \bfs_\bftheta(\tilde{\bfx}_t{+}\bm{\xi}^i, t, y^i) \right] + (1-\gamma)\bfs_\bftheta(\tilde{\bfx}_{t}, t, \varnothing) \right\} + \sigma_{t}\tilde{\eps}_{t} 
\end{aligned}
\end{equation}
Let $\bm{\Phi}_t$ be the compositional estimation at timestep $t$: 
\begin{equation}
    \bm{\Phi}_t =  \sum_{i=1}^n \frac{\bm{\mathcal{M}}^i}{\sum_{i=1}^n \bm{\mathcal{M}}^i} \otimes \bfs_\bftheta(\tilde{\bfx}_t{+}\bm{\xi}^i, t, y^i).
\end{equation}
Finally, we have 
\begin{equation}
    \tilde{\bfx}_{t{-}1} = \tilde{\alpha}_{t} \tilde{\bfx}_{t} + \tilde{\beta}_{t} \left[\gamma~\bm{\Phi}_t + (1-\gamma)\bfs_\bftheta(\tilde{\bfx}_{t}, t, \varnothing)\right] + \sigma_{t}\tilde{\eps}_{t}.
\end{equation}

\section{More analysis about Vision Guidance} 
\begin{figure}[!t]
    \centering
    \includegraphics[width=0.8\linewidth]{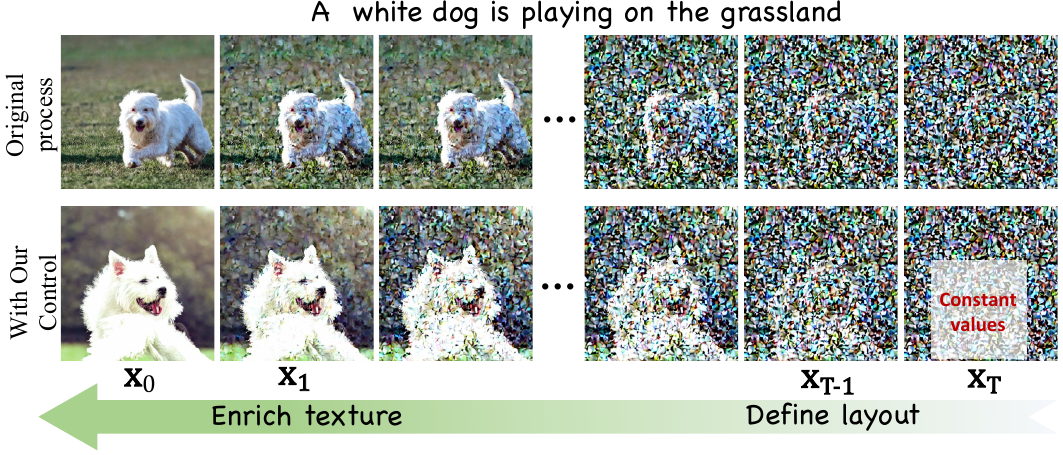}
    \caption{Top: general reverse diffusion process; Bottom: a reverse diffusion when adding small constant values to a region in the initial noise. The altered region defines the location and size of the synthesised object.}
    \label{vis_denosing}
\end{figure}
We first illustrate the effectiveness of the vision guidance mechanism from a simple experiment. A common knowledge of diffusion models is that the spatial layout of a synthesised image is established during the early reverse process. This phenomenon triggers our thoughts to conduct a preliminary exploration: for each channel, we manually add a small value (a constant version of vision guidance) to a local region within the initial noise, such as adding white colour feature values for a white dog. The lower part of Fig.~\ref{vis_denosing} illustrates the denoising process for this altered initial noise. Intriguingly, we discovered that the region we altered in the initial noise corresponded with where an object, mentioned in the text caption, appeared in the synthesised image. This finding indicates that altering the mean value of a local area in the initial noise can effectively guide the direction of the denoising process.

We further clarify the effectiveness of vision guidance from a distribution transfer perspective. The traditional diffusion denoising process can be seen as transferring the distribution, guided by text or other conditions, from a Gaussian distribution to a dataset-style distribution. In each denoising step, Unet estimates the noise from $X_t$. Subsequently, we caculate the mean of the distribution at the previous step $t-1$ and sample $X_{t-1}$. From the main text, vision guidance implies the layout by suppressing the tendency of object generation in the background area while enhancing it in the target area. Within $\xi$, we define $\delta$ as calculated features associated with the object text in the noise feature space, increasing attention and thus influencing the tendency. 
Adding a constant tensor $\delta$ to the $i_{th}$ point in $X_t^i$, denoted as $X_t^i + \delta$, implies sampling a point from the distribution $\mathcal{N}(u_t+ \delta, \sigma_t)$, indicating a preference for generating approximate object features. Conversely, subtracting a constant tensor $\delta$ from the $i_{th}$ point in $X_t^i$, denoted as $X_t^i - \delta$, suuggests sampling a point from the distribution $\mathcal{N}(u_t- \delta, \sigma_t)$, reflecting a preference for not generating approximate object features. The layout ability arises from points in different areas being drawn from distributions with varying tendencies to generate objects. To maintain consistency across distributions, we introduce a small coefficient in  $\delta$. Based on the above, before each normal denoising, vision guidance slightly advances the distribution that conforms to the object prompt to one that also conforms to the given layout. After several times, there will be a high probability of generating objects in the target area, and the pre-trained model can generate a result consistent with the layout from the current noise gradually.

From the perspective of attention mechanisms, a more intuitive explanation is that by incorporating features more relevant to the object text in specific areas, we enhance the similarity coefficient between these area features and the object text features. This, in turn, improves the accuracy of generating objects in these areas, based on probabilities.

\section{Dataset Construction} 
\begin{figure}[!h]
    \centering
    \includegraphics[width = 0.8\linewidth]{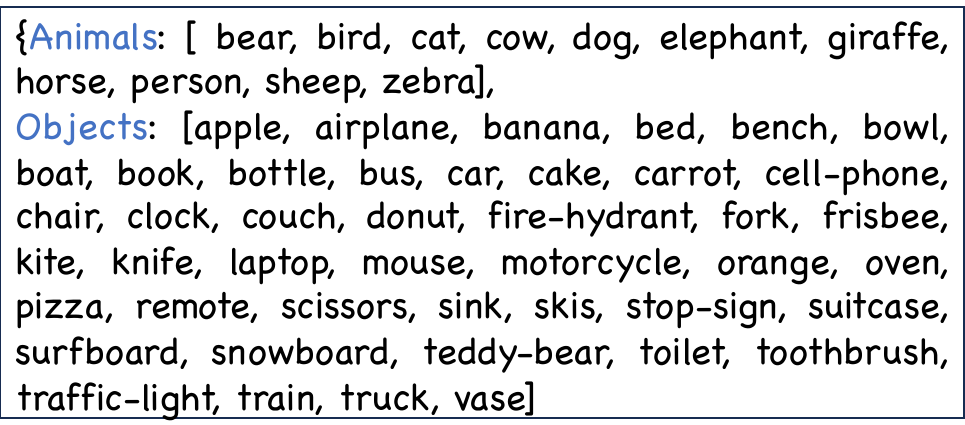}
    \caption{The constructed dataset contains 55 categories in total.}
    \label{category information}
\end{figure}

We construct a dataset comprising 1,134 global captions along with corresponding bounding boxes, object categories, and instance masks sourced from the MS COCO dataset. The dataset encompasses 55 categories in total (including 11 animal classes and 44 object classes) for spatially conditional text-to-image synthesis, which is summarised in 
In this dataset, for the layers from $[1, n-1]$, we configured the layered captions in the form of ``a $<\mathrm{category~name}>$", whereas, for the final layer $n$, the scene descriptions or other object categories are extracted using LLM from the global captions. Thanks to the exceptional capability of LLM in few-shot learning, we only provide a few examples, \eg, {``A bear is running the grassland", the description of the scene is ``The beautiful grassland"}, and the LLM effectively extracted the context and produced a suitable response. Notably, LLM is simply used for batch processing the global captions in the dataset. In practical applications, our method supports user-defined scene descriptions or provides a default description if not specified.
Fig.~\ref{category information}.

\section{More Visualisation Results}
\begin{figure*}[!h]
    \centering
    \includegraphics[width = 1.0\linewidth]{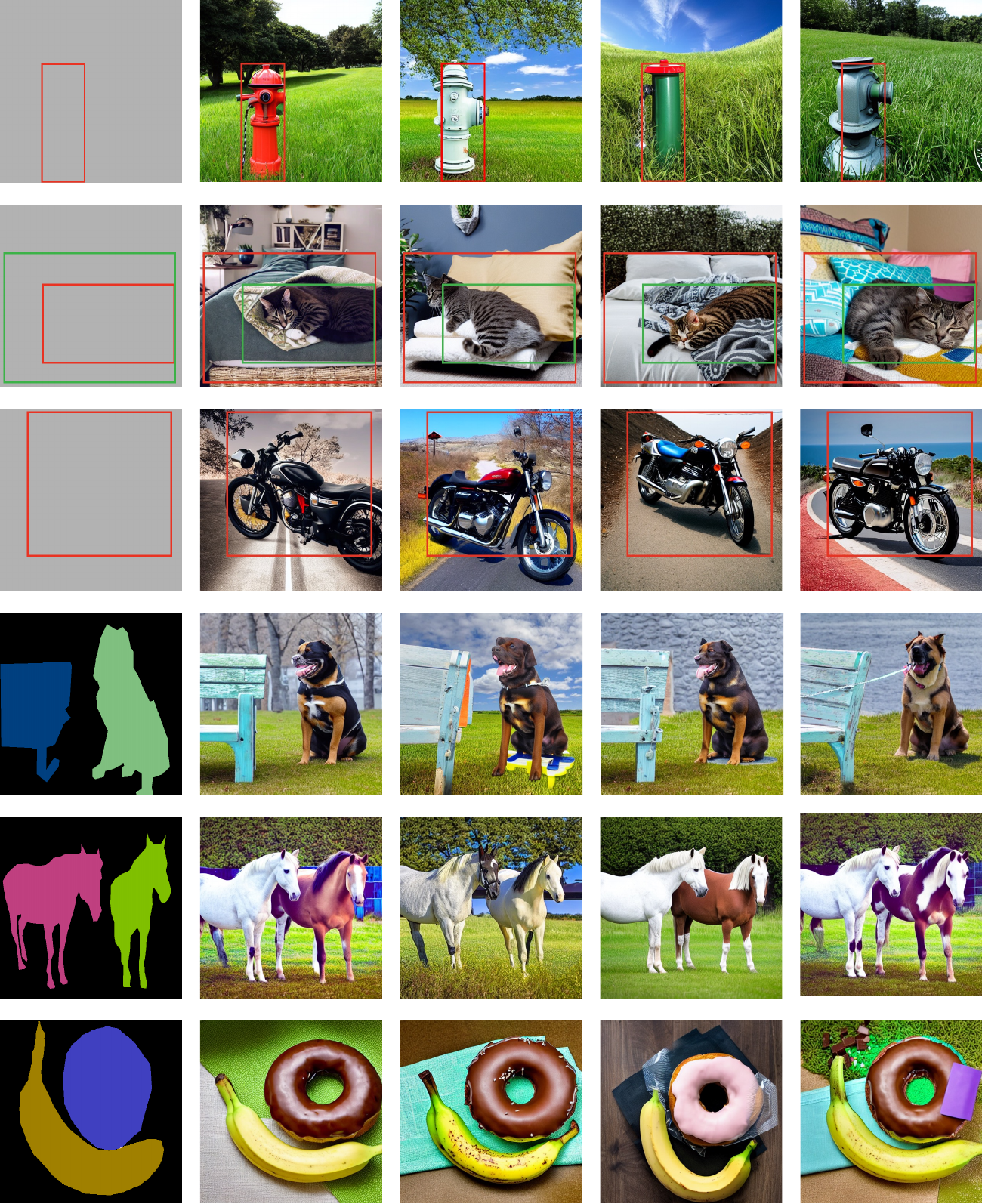}
    \caption{More visualisation results are presented using our method. Columns 2-5 showcase the outcomes with various random seeds.}
    \label{more visual results}
\end{figure*}

\begin{figure*}[!h]
    \centering
    \includegraphics[width = 1.0\linewidth]{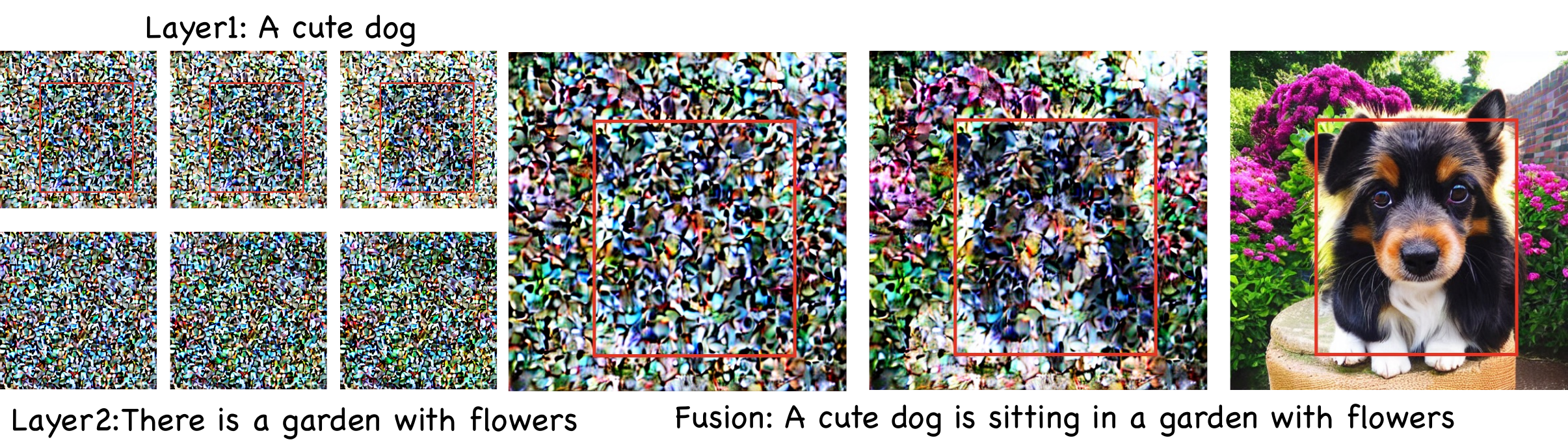}
    \caption{The visualisation of intermediate results of each layer.}
    \label{inter-results}
\end{figure*}

In Fig.\ref{inter-results}, we display the intermediate results for each layer. Meanwhile, Fig.\ref{more visual results} shows additional visualisation results obtained with different random seeds. 
\vspace{-0.3cm}

\end{document}